\documentclass[runningheads]{llncs}

\usepackage{accv}

\usepackage{multirow}
\usepackage{accvabbrv}

\usepackage{graphicx}
\usepackage{booktabs}
\usepackage{colortbl}  
\usepackage{xcolor}
\usepackage{array} 
\usepackage[accsupp]{axessibility}  

\usepackage[pagebackref,breaklinks,colorlinks,citecolor=accvblue]{hyperref}

\usepackage{orcidlink}

\begin{document}

\title{DiffLoss: unleashing diffusion model as constraint for training image restoration network} 

\titlerunning{Abbreviated paper title}
\author{Jiangtong Tan \and
Feng Zhao}

\authorrunning{J. Tan, F. Zhao.}

\institute{University of Science and Technology of China, China \\
\email{jttan@mail.ustc.edu.cn}, \email{fzhao956@ustc.edu.cn}}



\maketitle

\begin{abstract}
Image restoration aims to enhance low quality images, producing high quality images that exhibit natural visual characteristics and fine semantic attributes. Recently, the diffusion model has emerged as a powerful technique for image generation, and it has been explicitly employed as a backbone in image restoration tasks, yielding excellent results. However, it suffers from the drawbacks of slow inference speed and large model parameters due to its intrinsic characteristics.
In this paper, we introduce a new perspective that implicitly leverages the diffusion model to assist the training of image restoration network, called  DiffLoss, which drives the restoration results to be optimized for naturalness and semantic-aware visual effect.
To achieve this, we utilize the mode coverage capability of the diffusion model to approximate the distribution of natural images and explore its ability to capture image semantic attributes. On the one hand, we extract intermediate noise to leverage its modeling capability of the distribution of natural images, which serves as a naturalness-oriented optimization space. On the other hand, we utilize the bottleneck features of diffusion model to harness its semantic attributes serving as a constraint on  semantic level. By combining these two designs, the overall loss function is able to improve the perceptual quality of image restoration, resulting in visually pleasing and semantically enhanced outcomes.
To validate the effectiveness of our method, we conduct experiments on various common image restoration tasks and benchmarks. Extensive experimental results demonstrate that our approach enhances the visual quality and semantic perception of the restoration network.

\end{abstract}

\section{Introduction}
\label{sec:intro}
\begin{figure}[tb]
  \centering
  \includegraphics[width=0.9\linewidth]{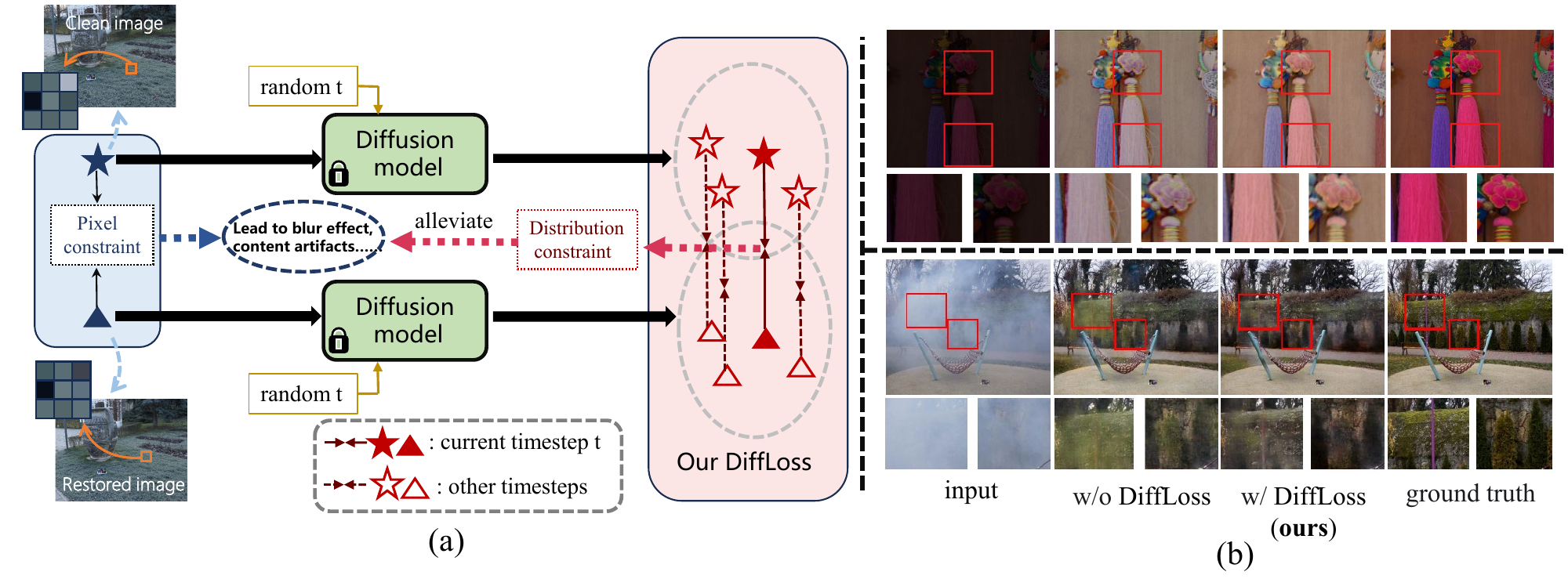}
  \caption{(a): Illustration of the effect of our method to convert pixel constraint into distribution constraint.(b): Visual comparison between baselines with and without our DiffLoss. The top is produced with IAT\cite{cui2022you} on  LOL dataset\cite{wei2018deep} and the bottom is produced with MSBDN\cite{dong2020multi} on NH-HAZE dataset~\cite{ancuti2020nh}. Previous loss is limited to the pixel level and suffers from unnaturalness problem, with color shift and content artifacts. Our DiffLoss leverages the powerful modeling capability of the diffusion model on the distribution of natural images, resulting in more natural outcomes in the image restoration process.  Need to mention that DiffLoss is an optimization strategy. The improvement should be compared with the baseline method, instead of other restoration methods or the ground-truth images.
  }
  \label{fig:Tissue}
\end{figure}

In complex imaging environments, the quality of imaging often suffers from unpredictable degradations, such as low-light conditions, heterogeneous media, leading to information loss both in content and color and unnatural visual effects. Besides, these degradations also causes adverse effects on high-level computer vision tasks such as object detection~\cite{chen2018domain,li2018end} and scene understanding~\cite{sakaridis2018semantic,sakaridis2018model}. Image restoration aims to solve these issues by recovering high quality images from degradations to achieve better natural visual quality and fine semantic attributes. This task has been a longstanding and challenging problem, drawing the attention of researchers. 
Previous traditional methods utilized the inherent and general statistical property of degraded images to design algorithms with prior knowledge for image restoration\cite{berman2016non,he2010single,fattal2014dehazing,fattal2008single}. 
However, these hand-crafted image priors are drawn from specific observations with limited robustness, which may not be reliable for modeling the intrinsic characteristics of degraded images. With the development of deep learning, image restoration has made marvelous progress, where most of them rely on designing advanced architectures to learn degradation to clean mappings \cite{li2017aod,liu2019griddehazenet,dong2020multi,wu2021contrastive, guo2022image,yufrequency}. 
However, due to the severe degradation and limited model capacity, the naturalness of the restored results is restricted by color and texture distortions, as shown in \cref{fig:Tissue}.
In addition to further improving image visual quality,  preserving semantic attributes is also significant for image restoration. Existing methods try to solve the problem by either involving multi-stage training or utilizing complex network structures\cite{kim2021quality, yang2022self, yang2023visual}, which result in inconvenience during implementation.
With the emergence of the diffusion model, it has shown remarkable performance as a backbone in image restoration tasks with naturalness and semantic-aware visual results\cite{choi2021ilvr, saharia2022image, whang2022deblurring, lugmayr2022repaint, kawar2022denoising, wang2022zero}.
However, these methods attempt to integrate diffusion models into the network architecture that has the limit of slow inference speed and accounting huge memory cost when deploying.

In this study, to address these issues, instead of creating new network architectures, we empower existing network frameworks with powerful prior on natural image distribution modeling and high-level semantic space.
Inspiringly, the diffusion model has exhibited strong natural image generation capability~\cite{dhariwal2021diffusion} and possible semantic potential \cite{baranchuk2021label}, which motivate us to exploit the "implicit" usage of the diffusion model and leverage it as an optimization prior to improve naturalness and semantic attributes for image restoration.


\begin{figure}[tb]
	\centering
     \begin{subfigure}{0.48\linewidth}
    \centering
    \includegraphics[width=0.90\linewidth]{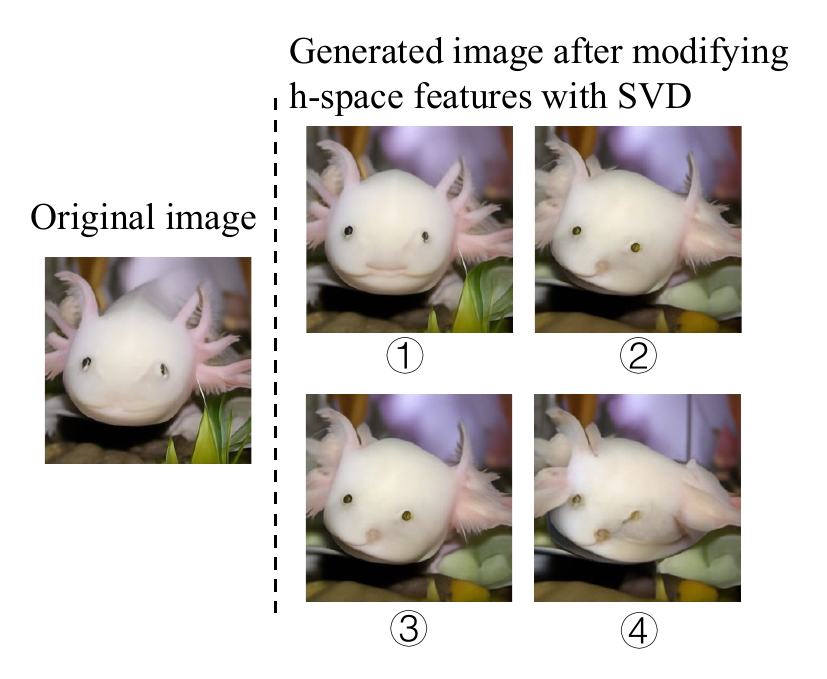}
    \caption{Illustration of images after modifying diffusion model's h-space features with SVD. From {\textcircled{\tiny{1}}} to {\textcircled{\tiny{4}}},  the h-space perturbation increases.}
    \label{fig:ha}    
    \end{subfigure}
    \begin{subfigure}{0.48\linewidth}
    \centering
    \includegraphics[width=0.90\linewidth]{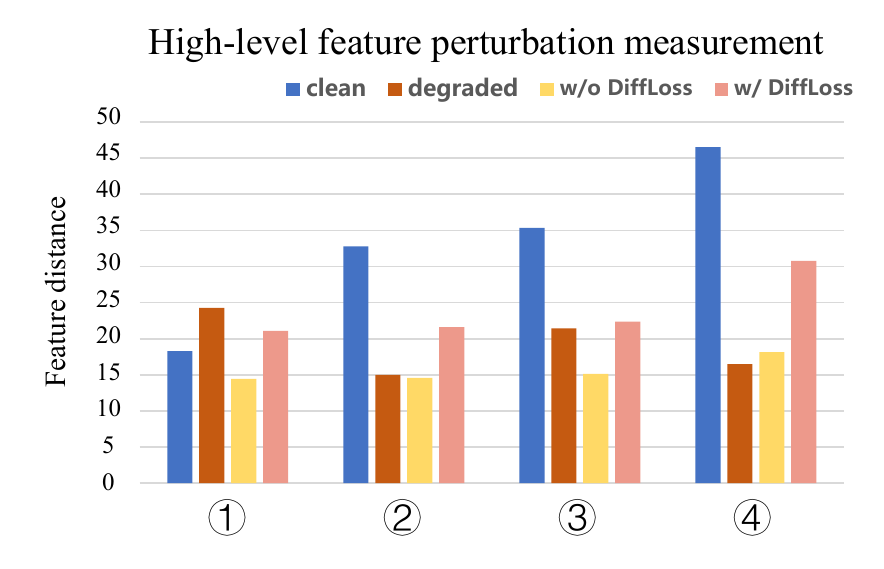}
    \caption{Histogram of high-level feature distance of results on low-light image enhancement with synthesized data from ImageNet dataset.}
    \label{fig:hb}
    \end{subfigure}
	\caption{As h-space changes, the image gradually loses its original semantics completely, from {\textcircled{\small{1}}} to {\textcircled{\small{4}}} in (a). We also employed the output features of the ResNet50 network to measure the distance of high-level features in the images with different types of degradation to show the change of semantic attributes, as depicted in the histogram (b). With h-space perturbation increase, clean images and restored images with DiffLoss exhibit systematic variations in semantic attributes, while the degraded images and restored images without DiffLoss show minimal changes, which means degradations undermine the semantic attributes of images, while our DiffLoss can restore this. Note that diffusion model and ResNet50 are both trained on ImageNet dataset.}
	\label{fig:hTissue}
\end{figure}

In this paper, we propose a new perspective to resort these issues by introducing a naturalness-oriented and semantic-aware optimization mechanism by using diffusion model, dubbed DiffLoss, which has two parts. (1) \textbf{Natural image distribution prior.} Drawing inspiration from the diffusion model's remarkable capacity to cover distributions in natural image generation, we leverage the Markov chain sampling characteristic of the diffusion model to project the restored results of existing networks into the noise sampling space and utilize it as a constraint to enhance the natural visual representation of images. Its improvement for natural visual performance can be observed in \cref{fig:Tissue}.  (2) \textbf{High-level semantic space prior.} The bottleneck feature of diffusion models, also dubbed h-space feature, is verified to be natural high-level semantic space in \cite{zhang2023unsupervised, haas2023discovering, jeong2024training} and is shown in \cref{fig:hTissue}. For both clear images and images with different degradations, we first add noise to them and then input them into the diffusion model. During the generation, we perform SVD decomposition on the feature in h-space and perturb its principal component. As the perturbation changes, the semantic appearance in the generated images also change. We extract the high-level features of these images in a ResNet50 that pre-trained on ImageNet for classification and measure the L2 distance of the perturbed image features from original image features. We find the perturbation in h-space can disrupt high-level features of clean images and restored images with our method. While for degraded images and restored images without our method, the disturbance is slightly, which implies degradations corrupt the semantic attributes of the images and DiffLoss can act as a constraint for semantic recovery in image restoration tasks.


For implementation, we employ the unconditional diffusion model pre-trained on ImageNet dataset as optimization prior to exploit its immense clean data distribution. 
The diffusion model takes the restored image and ground truth as inputs and projects them into the distribution sampling space and semantic space by extract their sampling noise distribution and bottleneck feature. Within these spaces, the empowered restoration network can get more natural result and preserve more semantic attributes by pulling the intermediate sampling noise distribution and bottleneck feature of the restored image closer to that of the corresponding clear image. 
Different from previous methods that focus on dedicated model design or directly employ diffusion model as the restoration model, our DiffLoss works as a general and novel auxiliary training mechanism, which can endow existing restoration methods with both more natural and semantic-aware results with this effective training strategy.
Additionally, it's especially beneficial for empowering parameter-limited models as it involves naturalness prior. We also compared it with other loss functions that assist restoration networks, demonstrating excellent performance. Our method also involves no additional computations in the inference stage and easy for implementation.
We verify the effectiveness of our method on substantial common image restoration tasks, including image dehazing, image deraining, and low-light image enhancement. 

\vspace{0.3cm}
Overall, our contributions can be summarized as follows:
\begin{itemize}
\item We introduce the naturalness and semantic-aware modeling paradigm into restoration network by embedding the diffusion model as an auxiliary training mechanism, which has not been explored before. This training strategy alleviates the low quality issue caused by unnatural visual quality and lack of semantic attributes in existing methods.
\vspace{0.1cm}
\item Specifically, our approach leverages the fixed diffusion model to enable the extraction of intermediate sampling noise and semantic information, and it yields more natural and semantic-aware restoration when the DiffLoss is minimized.
\vspace{0.1cm}
\item Extensive experiments demonstrate that our DiffLoss empowers existing restoration methods compared with other loss functions, helps training efficient models and improves the classification ability on data with varying of degradation without involving additional computations in the inference stage at all.
\end{itemize}

\section{Related work}
\subsection{Visual and semantic improvement for image restoration}
Image restoration refers to the process of improving the quality of a degraded or damaged image by removing various types of degradations. With the advent of deep learning, numerous methods leveraging deep neural networks for image restoration have emerged, such as low-light enhancement\cite{wei1808deep, zhang2021beyond, zhang2019kindling, zhu2020zero}, super-resolution\cite{ledig2017photo, lim2017enhanced, zhang2018image, zhang2018residual}, dehazing\cite{berman2016non, chen2019pms, fattal2008single, he2010single, li2020zero, li2017aod, zhang2018densely}, deraining\cite{fu2017removing, li2018recurrent, luo2015removing, qian2018attentive, ren2019progressive}, deblurring\cite{chakrabarti2016neural, cho2021rethinking, gao2019dynamic, kupyn2018deblurgan, ren2021deblurring}. However, the severe degradation limits the naturalness of restored results, causing color and texture distortions. Moreover, it is proved that solely relying on visual quality metrics during the restoration process without considering the semantic aspects of quality will negatively impacted performance in downstream tasks. Some efforts \cite{kim2021quality, yang2022self, yang2023visual} have attempted to improve the semantic attributes of restored network output images through time-consuming training strategies and complex networks, but they often result in inconvenience and redundancy in practical applications.

\subsection{Diffusion models for image restoration}
Existing diffusion models used in image restoration tasks can be roughly categorized into two classes: conditional diffusion and unconditional diffusion. 

Conditional DDPMs \cite{choi2021ilvr, saharia2022image, whang2022deblurring, rombach2022high, ozdenizci2022restoring, kawar2022denoising, wei2023raindiffusion} are usually combined with specific image restoration task, such as image super-resolution~\cite{saharia2022image}, image deblurring~\cite{whang2022deblurring}, and image deraining~\cite{wei2023raindiffusion}. 
SR3 trains a conditional diffusion model for image super-resolution with the low-resolution images as condition.
Whang et al.~\cite{whang2022deblurring} proposed the "predict and refine" strategy and learned the residual with conditional diffusion model in image deblurring task.
RainDiffusion~\cite{wei2023raindiffusion} combines cycle-consistent architecture with diffusion model to achieve unsupervised image deraining.

Unconditional DDPMs~\cite{lugmayr2022repaint, kawar2022denoising, wang2022zero, songpseudoinverse, mei2022bi} are usually integrated into general image restoration task. 
For example, RePaint~\cite{lugmayr2022repaint} solves inpainting problem by employing unconditional diffusion process in the unmasked region and reverse back to solve boundary inconsistency.    
DDRM~\cite{kawar2022denoising} uses SVD to decompose the degradation operators and embeds unconditional diffusion model into unsupervised posterior sampling method to solve various linear inverse problems.
DDNM~\cite{wang2022zero} applies range-null space decomposition to degraded images and refines only the null-space contents during the reverse process to yield diverse results.

However, these methods aim to incorporate diffusion models as backbone, which suffer from the drawbacks of slow inference speed and significant memory consumption during deployment. We circumvent these limitations from a new perspective by exploring diffusion model as an auxiliary training mechanism to empower the learning capability of existing image restoration networks, without involving additional computations in the inference stage.

\begin{figure}[tb]
  \centering
  \includegraphics[width=1\linewidth]{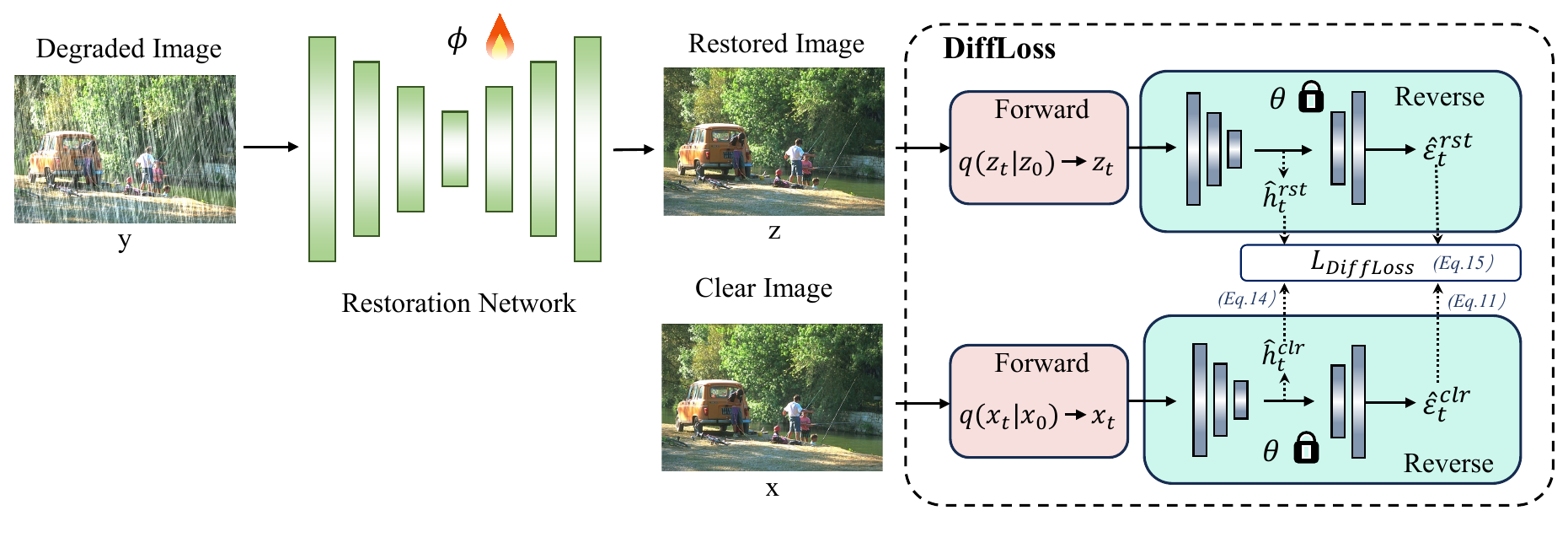}
  \caption{Overview structure of our method. The parameter of DiffLoss is frozen during training stage. For any existing restoration network, we train it with the aid of our DiffLoss to achieve higher natural visual and semantic performance. More implementation details of DiffLoss can be found in \cref{fig:ppl2}. During inference stage, we only have the optimized restoration network, without involving the DiffLoss.
  }
  \label{fig:ppl}
\end{figure}


\section{Methodology}
\label{sec:blind}
In this section, we first briefly introduce the denoising diffusion
probabilistic models, followed by a detailed presentation of the DiffLoss we  propose.

\subsection{Denoising Diffusion Probabilistic Models}
DDPM is a latent variable model specified by a T-step Markov chain, which approximates a data distribution $q(x)$ with a model $f_{\theta}(\cdot)$. It contains two processes: the forward diffusion process and the reverse denoise process.

{\bf The forward diffusion process.}
The forward diffusion process starts from a clean data sample $x_0$ and repeatedly injects Gaussian noise according to the transition kernel $q(x_t|x_{t-1})$ as follows:
\begin{equation}
	\small
	\label{1}
	q(x_t|x_{t-1}) = N(x_t;\sqrt{\alpha_t}x_{t-1},(1-\alpha_t)I),  \\
\end{equation}
where $\alpha_t$ can be learned by reparameterization~\cite{kingma2013auto} or held constant as hyper-parameters, controlling the variance of noise added at each step.
From the Gaussian diffusion process, we can derivate closed-form expressions for the marginal distribution $q(x_t|x_0)$ and the reverse diffusion step $q(x_{t-1}|x_t,x_0)$ as follows:
\begin{align}
\label{2}
    q(x_t|x_0) = N(x_t;\sqrt{\bar{\alpha} _t}x_0,(1-\bar{\alpha}_t)I), \\
\label{3}
    q(x_{t-1}|x_t,x_0) = N(x_{t-1};\tilde{\mu}_t(x_t,x_0),\tilde{\beta}_tI),
\end{align}
where $\tilde{\mu}_t(x_t,x_0) := \frac{\sqrt{\bar{\alpha}_{t-1}}(1-\alpha_t)}{1-\bar{\alpha}_t}x_0 +  \frac{\sqrt{\alpha_t}(1-\bar{\alpha}_{t-1})}{1-\bar{\alpha}_t}x_t$, $\tilde{\beta}_t:= \frac{1-\bar{\alpha}_{t-1}}{1-\bar{\alpha}_t}(1-\alpha_t)$, and $\bar{\alpha_t} := {\textstyle \prod_{s=1}^{t}} \alpha_s$.

Note that the above-defined forward diffusion formulation has no learnable parameters, and the reverse diffusion step cannot be applied due to having no access to $x_0$ in the inference stage. Therefore, we further introduce the learnable reverse denoise process for estimating $x_0$ from $x_T$.

{\bf The reverse denoise process.}
The DDPM is trained to reverse the process in Equation \ref{1} by learning the denoise network $f_\theta$ in the reverse process. Specifically, the denoise network estimates $f_\theta(x_t,t)$ to replace $x_0$ in Equation \ref{3}.
Note that $f_\theta(x_t,t)$ directly predicts the Gaussian noise $\varepsilon$, instead of $x_0$. While, the estimated $\varepsilon$ deterministicly corresponds to $\hat{x}_0$ via Equation \ref{2}.

\begin{equation}
	\small
	\label{4}
	\begin{aligned}
    p_\theta(x_{t-1}|x_t) &= q(x_{t-1}|x_t,f_\theta(x_t,t))\\
    &= N(x_{t-1};\mu _\theta(x_t,t),{\textstyle \sum_{\theta}^{}} (x_t,t)).\\
    \end{aligned}
\end{equation}
\begin{equation}
	\small
	\label{5}
	\mu _\theta(x_t,t) = \tilde{\mu_t}(x_t,x_0),  {\textstyle \sum_{\theta}^{}} (x_t,t) = \tilde{\beta_t}I.
\end{equation}

Similarly, the mean and variance in the reverse Gaussian distribution \ref{4} can be determined by replacing $x_0$ in $\tilde{\mu}_t(x_t,x_0)$ and $\tilde{\beta}_t$ with the learned $\hat{x}_0$

{\bf Training objective and sampling process.}
As mentioned above, $f_\theta(x_t,t)$ is trained to approach the Gaussian noise $\varepsilon$. Thus the final training objective is:
\begin{align}
\label{6}
    L=E_{t,x_0,\varepsilon}\left \| \varepsilon -f_\theta(x_t,t) \right \|_1.
\end{align}

The sampling process in the inference stage is done by running the reverse process. Starting from a pure Gaussian noise $x_T$, we iteratively apply the reverse denoise transition $p_\theta(x_{t-1}|x_t)$ T times, and finally get the clear output $x_0$.
\begin{figure}[tb]
  \centering
  \includegraphics[width=1\linewidth]{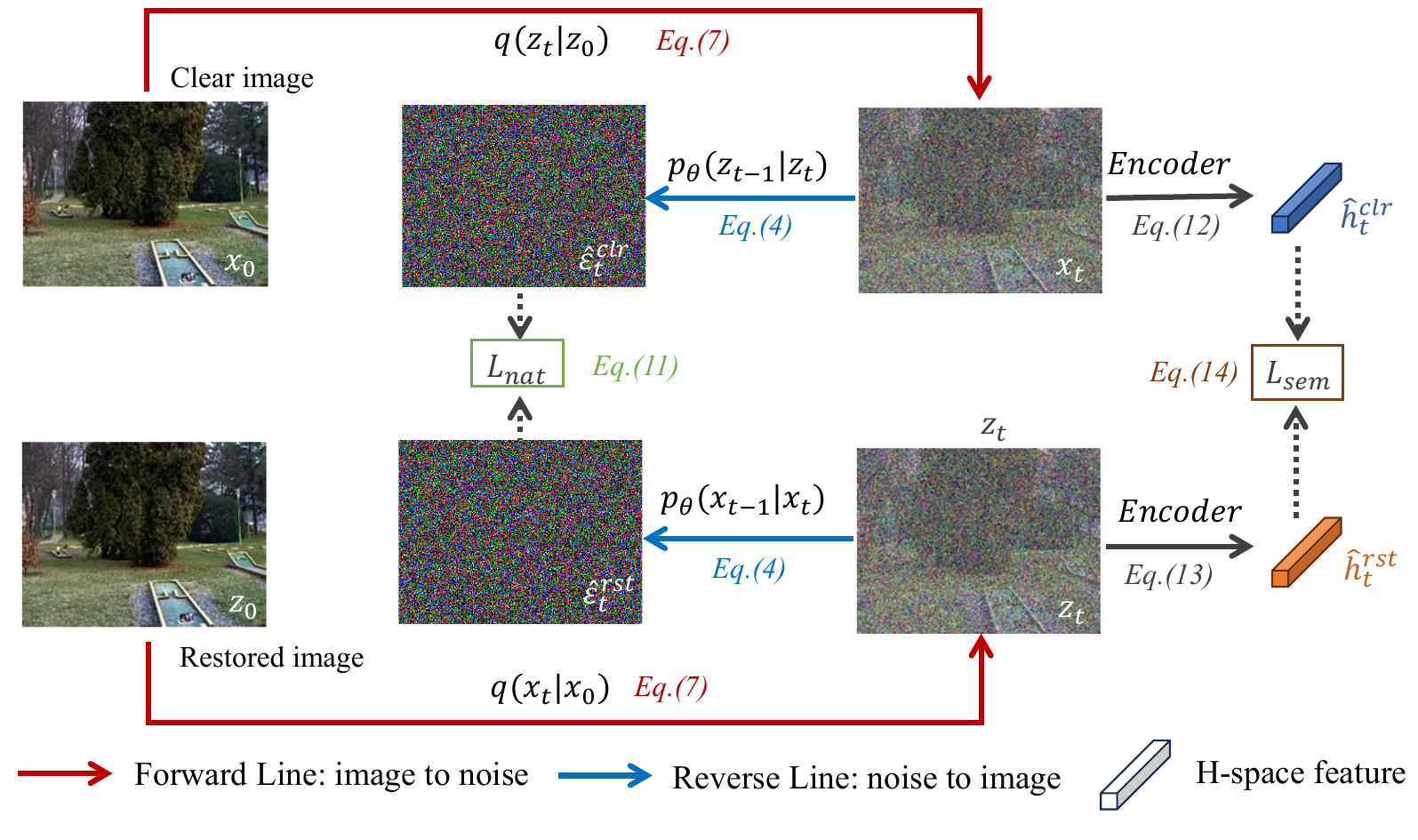}
  \caption{Detailed design of DiffLoss. We devise our DiffLoss with $t$ step forward process and one step reverse process. The $t$ step forward process can be directly achieved with \cref{7}. Then we projecting these noisy images into intermediate noise with \cref{4} after fed into the denoising UNet. We also get h-space vector from bottleneck of the UNet,which contains semantic information, as described in \cref{12} and \cref{13}. The DiffLoss is designed to pull the output of the denoising UNet and the bottleneck feature closer. 
  }
  \label{fig:ppl2}
\end{figure}
\subsection{Overview structure}
As presented in \cref{fig:ppl}, in our main setting, given the degraded image $y$, its corresponding clear image $x$ , any existing restoration network $g_{\phi}(\cdot)$, and pre-trained diffusion model $f_{\theta}(\cdot)$ (with fixed parameter $\theta$), we first get the restored output $z = g_\phi(y)$ from the restoration network. Then DiffLoss works as auxiliary training mechanism providing naturalness modeling ability and for restoration networks.
\subsection{Detailed design of DiffLoss}
Originally, the forward diffusion process translates a clean data sample $x_0$ into Gaussian noise $\varepsilon$ by gradually adding Gaussian noise to $x_0$ in a parameter-free manner. The reverse denoise process is trained to sample from Gaussian noise $\varepsilon$ to generate clean images via gradually removing noise with the denoise network $f_\theta$. However, both of these two processes have asymmetric input-output pairs and thousands of iterative steps. These properties are not suitable for direct loss design. For example, direct applying the reverse denoise process $T$ times is time-consuming and impacts the backpropagation of gradients. Besides, since the DiffLoss takes the restored result $z$ as input, it should start from forward process, and connect to the reverse process in a proper way. Because single forward process is parameter-free and can be used as a learnable loss.

To this end, we redesign diffusion model delicately. We employ $t$ step forward diffusion process and one step reverse denoise step to minimize time-consuming as well as get symmetric image-image input-output pairs. Specifically, as shown in  \cref{fig:ppl} and \cref{fig:ppl2}, we integrate the forward diffusion process and reverse denoise process. In the forward diffusion process, we get the intermediate noisy image $x_t$ via $q(x_t|x_0, \varepsilon)$. Note that $t$ is obtained through uniform distribution sampling, which is expressed as \cref{7}. And \cref{8} provides the way of reconstructing $x_0$ back:

\begin{align}
	\label{7}
	x_t = \sqrt{\bar{\alpha}_t}x_0 + \sqrt{(1-\bar{\alpha}_t)}\varepsilon . \\
 	\label{8}
	x_0 = \frac{1}{\sqrt{\bar{\alpha}_t}}x_t - \sqrt{(\frac{1}{\bar{\alpha}_t} - 1)}\varepsilon.
\end{align}

Then comes the reverse denoise process. We need to implement  \cref{8} in a learnable way. Accordingly, we devise a way by respectively replacing $\varepsilon$ with the diffusion model learned ones, as shown in \cref{fig:ppl}.

Starting from clear image with added noise $x_t$, we get the pseudo Gaussian noise $\hat{\varepsilon}^{clr}_{t} = f_\theta(x_t, t)$. Then, the pseudo Gaussian noise map $\hat{\varepsilon}^{clr}_{t} $ is employed to replace $\varepsilon$ in \cref{8}. The inverse way $q(\hat{x_0}|x_t, \hat{\varepsilon}^{clr}_{t} )$ is expressed as follows:
\begin{align}
\label{9}
    \hat{x}_0 = \frac{1}{\sqrt{\bar{\alpha}_t}}x_t - \sqrt{(\frac{1}{\bar{\alpha}_t} - 1)}\hat{\varepsilon}^{clr}_{t} .
\end{align}

Similarly, the output results of the restoration network can undergo the same process to obtain the following equation.
\begin{align}
\label{10}
    \hat{z}_0 = \frac{1}{\sqrt{\bar{\alpha}_t}}z_t - \sqrt{(\frac{1}{\bar{\alpha}_t} - 1)}\hat{\varepsilon}^{rst}_{t} .
\end{align}




Note that we can also obtain $\hat{x}_{t-1}$ and $\hat{z}_{t-1}$ using \cref{4}. We choose to direct constrain on $\hat{\varepsilon}^{clr}_{t}$ and $\hat{\varepsilon}^{rst}_{t}$ without reconstructing back to $\hat{x}_0$ and $\hat{z}_0$ or using $\hat{x}_{t-1}$ and $\hat{z}_{t-1}$ as constraint, cause it has best performance and is shown in \cref{sec:abl}. The loss function can be expressed as follows:

\begin{align}
\label{11}
       L_{nat} &= \left \| \hat{\varepsilon}^{clr}_{t} - \hat{\varepsilon}^{rst}_{t} \right \|_2 .
\end{align}

In this way, the restoration networks can harness the generative capabilities of the diffusion model, as well as the sampling space's reflection of the natural attributes of images, to obtain restored results with more naturalness.

If we decompose $f_\theta(x_t, t)$ into an encoder $\mathcal{E}_\theta(\cdot)$ and a decoder $\mathcal{D}_\theta(\cdot)$ , then after obtaining the noisy version of clear image $\hat{x}_t$ and restored image $\hat{z}_t$, the form of semantic feature of them can be written as the following equations respectively:
\begin{align}
\label{12}
    \hat{h}_{t}^{clr} = \mathcal{E}_\theta(x_t, t),\\
\label{13}
    \hat{h}_{t}^{rst} = \mathcal{E}_\theta(z_t, t),
\end{align}
where $\hat{h}_{t}^{clr}$ and $\hat{h}_{t}^{rst}$ reflect bottleneck features in the middle layers of U-Net from the clear image and restored image, respectively. By reducing the gap between these two terms, we can preserve more semantic information in the restored image, which is expressed as follows:

\begin{align}
\label{14}
       L_{sem} &= \left \| \hat{h}_{t}^{clr} - \hat{h}_{t}^{rst} \right \|_2 .
\end{align}

Besides our newly employed DiffLoss, we preserve the traditional L2 loss between $z$ and $x$ for stable optimization.
In conclusion, our DiffLoss and total losses used in the training stage is expressed as follows:
\begin{flalign}
\label{15}
    L_{DiffLoss} &= L_{nat} + \lambda L_{sem}, \\
\label{16}
    L_{total} &= \left \| x - z \right \|_2 + \gamma L_{DiffLoss}.
\end{flalign}
where $\lambda$ and $\gamma$ are weight factors. We set the weight $\lambda = 0.01$ and $\gamma = 0.001$, which is discussed in \cref{sec:abl}. Besides, we also try adaptive weight which is correlated with timestep $t$. Both strategies perform similarly. 
By employing these two strategies, the DiffLoss has the potential to enhance existing restoration methods, yielding restored images that are not only more natural but also reserve more semantic information.

\section{Experiments}
In this section, we first introduce the datasets and implement details of our experiment. Then, we conduct experiments on existing restoration methods, including low light enhancement, image deraining and image dehazing, with and without our DiffLoss. To validate the effectiveness of our approach in low-level tasks, we conducted tests on the efficiency model. Additionally, we examined the performance of the model employing our method in the image classification task with degraded data. Experiments on several baselines and benchmarks demonstrate the effectiveness of our DiffLoss.

\begin{table*}[t]
 \small
    \centering
    \caption{Quantitative comparison on three datasets. $\uparrow$ indicates that the larger the value, the better the performance. $``*"$ refers to the efficient model of the task.}
    \resizebox{1\columnwidth}{!}{
    
    \begin{tabular}{c|c|c|c|c|c|c|c|c}
    \toprule
    \multicolumn{3}{c|}{Dehazing} & \multicolumn{3}{c|}{Deraining}  & \multicolumn{3}{c}{Low light enhancement}\\
    \cline{1-9}
    \multirow{2}{*}{Method} & 
    \multicolumn{2}{c|}{Dense-Haze} &\multirow{2}{*}{Method} &  \multicolumn{2}{c|}{Rain100H} & \multirow{2}{*}{Method} & \multicolumn{2}{c}{LOL} \\
    \cline{2-3}
    \cline{5-6}
    \cline{8-9}
      & PSNR$\uparrow$ & SSIM$\uparrow$  & &PSNR$\uparrow$ & SSIM$\uparrow$  & &PSNR$\uparrow$ & SSIM$\uparrow$  \\

    \cline{1-9}
    AOD-Net\cite{li2017aod} & 13.14 & 0.4144 & DerainNet\cite{fu2017clearing} & 14.92 & 0.5920 & EnlightenGAN\cite{jiang2021enlightengan} & 17.48 & 0.6510 \\
    FFA-Net\cite{qin2020ffa} & 14.39 & 0.4524 & RESCAN\cite{li2018recurrent} & 26.36 & 0.7860 & RetiNexNet\cite{wei1808deep} & 16.77 & 0.5620 \\
    AECR-Net\cite{wu2021contrastive} & 15.80 & 0.4660 & PreNet\cite{ren2019progressive} & 26.77 & 0.8580 & DRBN\cite{xie2023semi} & 19.55 & 0.7460 \\
    \cline{1-9}
    FSDGN\cite{yufrequency} & 14.34 & 0.4010 & RCD-Net\cite{wang2020model} & 17.11 & 0.4634 & DeepLPF\cite{moran2020deeplpf} & 18.44 & 0.7431 \\
    \rowcolor{lightgray!40} w/DiffLoss & 14.66 & 0.4064 & w/DiffLoss & 17.67 & 0.4710 & w/DiffLoss & 19.51 & 0.7437 \\
    \cline{1-9}
    TaylorFormer*\cite{qiu2023mb} & 15.02 & 0.5178 & EfDeRain*\cite{guo2021efficientderain} & 23.41 & 0.7524 & IAT*\cite{cui2022you} & 19.89 & 0.7371 \\
    \rowcolor{lightgray!40} w/DiffLoss & 15.19 & 0.5326 & w/DiffLoss & 24.54 & 0.7656 & w/DiffLoss & 20.11 & 0.7378 \\

    \bottomrule
  \end{tabular}}
  
  \label{tab:base}
\end{table*}

\subsection{Experiment Setup}
\noindent
{\bf Datasets.} 
We train and evaluate our models on both synthetic and real-world image restoration datasets, including low light enhancement, image deraining and image dehazing. For real-world challenging scenes, we adopt LOL dataset\cite{wei2018deep} for low light enhancement and Dense-Haze dataset\cite{Dense-Haze_2019} for image dehazing. For image deraining, we adopt Rain13K dataset\cite{chen2021hinet} for training and Rain100H dataset\cite{yang2017deep} for testing. 
Finally, we use CUB dataset\cite{wah2011caltech} for image classification task, and the degraded 
 CUB dataset is obtained through synthetic methods from \cite{hendrycks2019benchmarking} to simulate degraded conditions, such as fog, rain, and low-light scenarios.

\begin{figure}[t]
	\centering
     \begin{subfigure}{0.90\linewidth}
    \centering
    \includegraphics[width=0.95\linewidth]{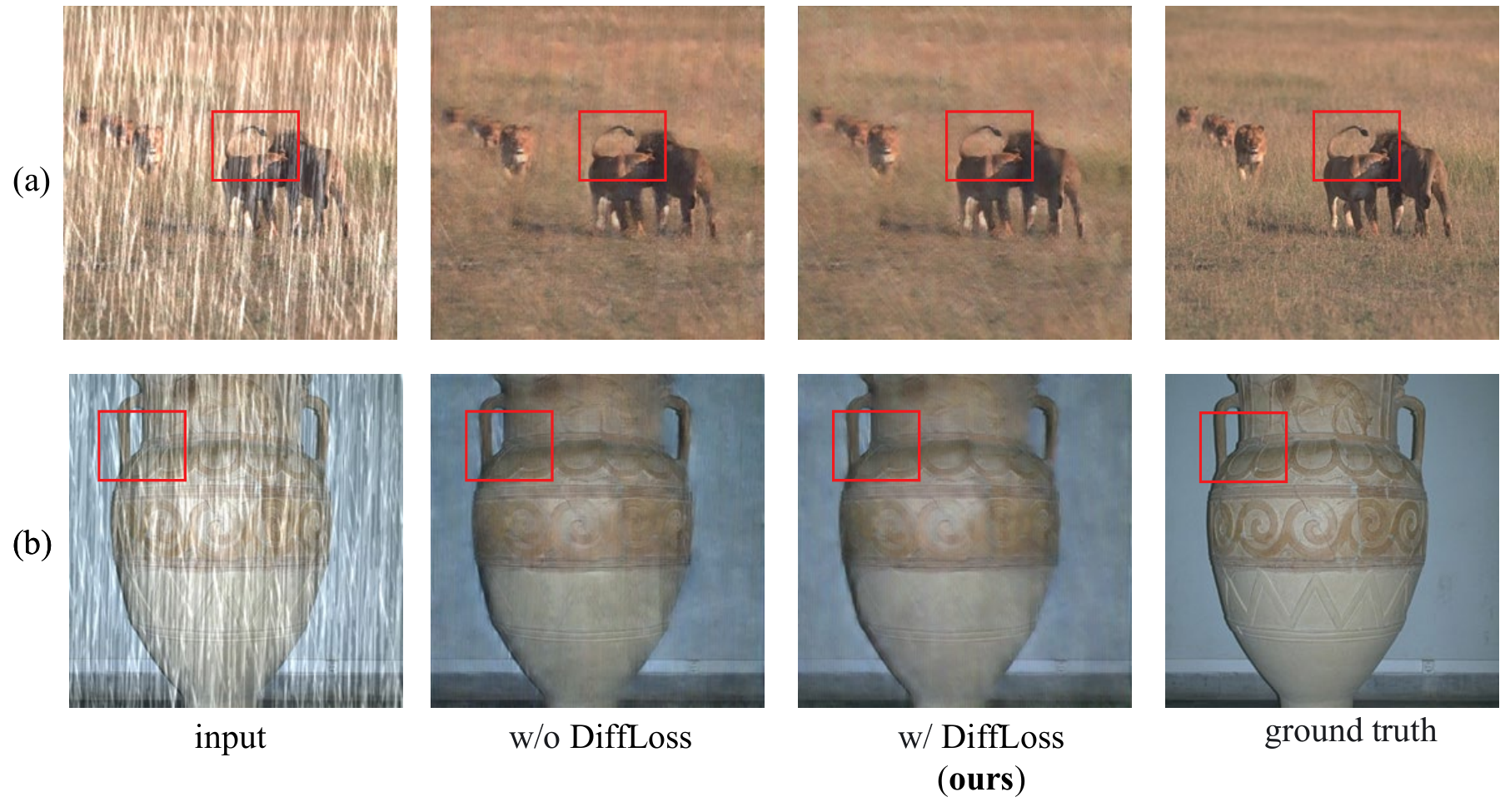}
    \label{fig:res1}    
    \end{subfigure}
    
    \begin{subfigure}{0.90\linewidth}
    \centering
    \includegraphics[width=0.95\linewidth]{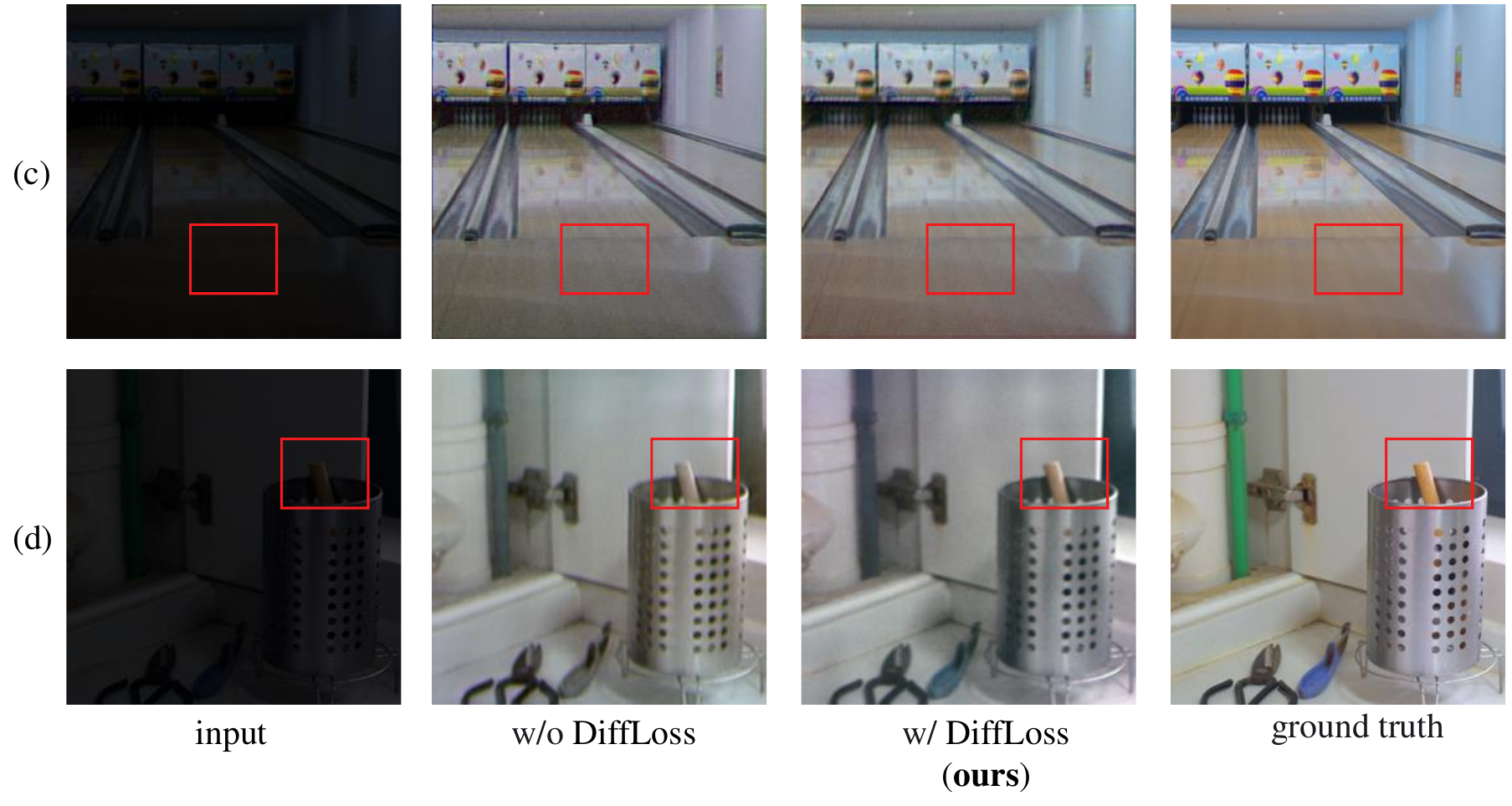}
    \label{fig:res2}
    \end{subfigure}

	\caption{Comparison of visual results on Rain100H and LOL datasets. (a):EfDeRain; (b): RCD-Net; (c): IAT; (d): DeepLPF. Please zoom in for best view.}
	\label{fig:res}
\end{figure}

\noindent
{\bf Comparison Baseline Methods.} 
We choose several classical and SOTA restoration methods as baselines, including IAT\cite{cui2022you}, DeepLPF\cite{moran2020deeplpf} for low light enhancement, EfDeRain\cite{guo2021efficientderain}, RCD-Net\cite{wang2020model} for image deraining, and  TaylorFormer\cite{qiu2023mb}, FSDGN~\cite{yufrequency} for image dehazing. We separately train these baselines with and without our DiffLoss, with the same settings and implementation details. 
We compare these two settings on the above baselines qualitatively and quantitatively.

\begin{figure}[t]
	\centering
    \includegraphics[width=1\linewidth]{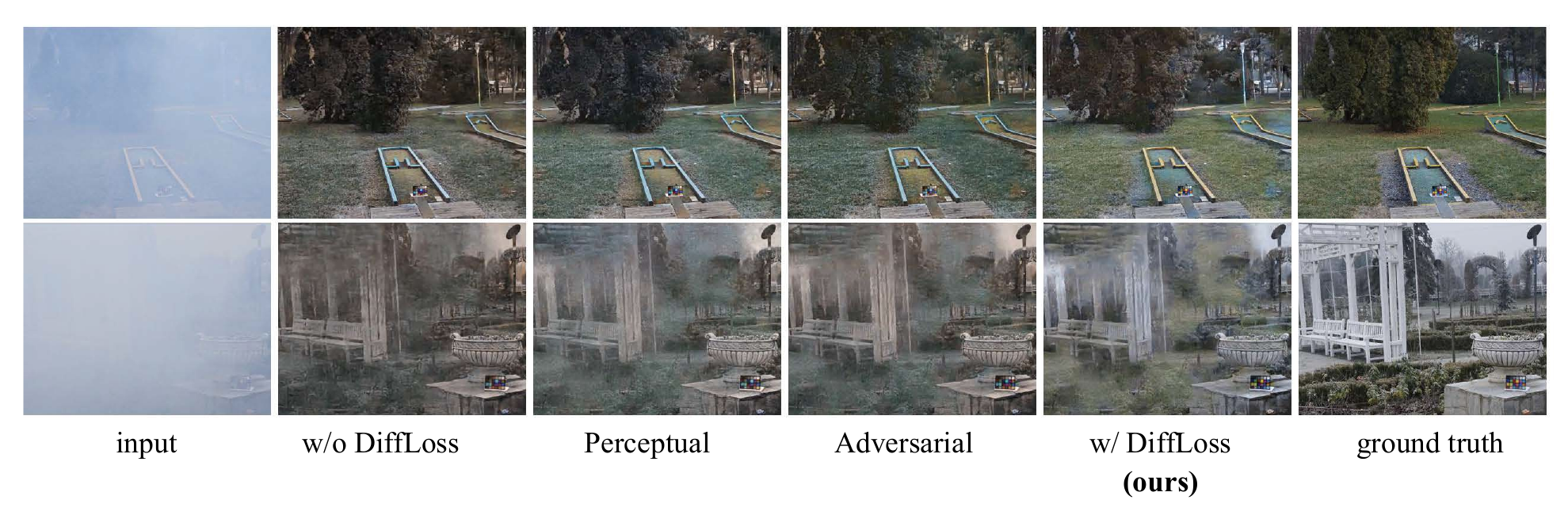}
	\caption{Comparison of visual results on Dense-Haze dataset. We also show the visual comparison between different loss functions and our method.}
	\label{fig:reshaze}
\end{figure}

\noindent
{\bf Implementation Details.}
We choose existing restoration networks listed above as backbone. For the DiffLoss, we adopt the unconditional diffusion model pre-trained on ImageNet dataset by~\cite{dhariwal2021diffusion}. Both the network architecture of restoration network and diffusion model need no modification. Besides, we fix the parameters of diffusion model. During the training stage, we use ADAM as the optimizers with the learning rate set to $1\times10^{-4}$. The batch and patch sizes are set to $4$ and $256\times256$, respectively. The parameter size of diffusion model is 552.81M. For image classification, we use VGG16\cite{simonyan2014very} and Resnet50\cite{he2016deep} pre-trained on clean CUB dataset as the recognition models to evaluate images restored by different methods. All the restoration models are trained with RTX 4090 GPU.

\noindent
{\bf Evaluation Metrics.} 
We evaluate our method on two different metrics: PSNR, SSIM~\cite{wang2004image}, LPIPS\cite{zhang2018unreasonable} and FID\cite{yu2021frechet} which are well-known image quality assessment indicators.

\subsection{Comparison with Baseline Methods on Several Benchmarks.}
\label{sect:figures}
First of all, need to note that our method improves the naturalness of restored results, instead of substantially removing more degradation. 

\noindent 
{\bf Comparison on real-world degradation images.} 
\cref{tab:base} compares the quantitative results of different methods on Dense-Haze, Rain100H and LOL datasets for image dehazing, image deraining and low light enhancement, respectively. On these datasets, Baselines with DiffLoss achieves better performance with most of metrics. The results on these challenging real-world datasets effectively demonstrate the advantages and effectiveness of our approach.

We also specifically utilized the efficient model, including TaylorFormer, EfDeRain and IAT, and the experiments revealed that our method enables the efficient model to achieve better image restoration. This is attributed to the inherent generative capability of the diffusion model, which helps the efficient model learn the distribution of natural images.

We also show the visual comparison with other method on real degradation images sampled from testing sets of Dense-Haze, Rain100H and LOL datasets. The visual results are presented in ~\cref{fig:res} and \cref{fig:reshaze}. The complex degradation distribution makes these datasets extremely challenging, and existing methods on these datasets usually suffer from unnaturalness problems. As shown in  ~\cref{fig:res} and \cref{fig:reshaze}, these baselines produce relatively pleasing results, however, artifacts and blurs still emerge. In contrast, after being empowered by the DiffLoss, they produce results more visually pleasing than the baseline results. We also present a comparison with different loss functions in \cref{fig:reshaze}, which is described in the following sections. For better visualization, we denote the obvious region with the red rectangular box.


\begin{table}[t]
\small
\centering
\caption{The results of Image Classification on CUB dataset among three different degradation. “Top-1 V” and “Top-1 R” refer to the Top-1 Accuracy ($\%$) on pre-trained VGG16\cite{simonyan2014very} and Resnet50\cite{he2016deep}, respectively.}
\begin{tabular}{c|c|c|c|c|c|c}
	\toprule
	\multirow{2}{*}{Config}  & \multicolumn{2}{c|}{dehazing}  & \multicolumn{2}{c|}{deraing} & \multicolumn{2}{c}{low light enhancement } \\
	\cline{2-7}
	
      & Top-1 V  & Top-1 R & Top-1 V  & Top-1 R & Top-1 V  & Top-1 R   \\
      \midrule
        w/o DiffLoss  & 37.5940  & 24.3966 & 68.6724 & 78.8966  & 15.2069 & 28.2069  \\
        \rowcolor{lightgray!40} w/ DiffLoss  & 46.7105 & 37.5906 & 70.0517 & 79.5690  & 35.7241 & 53.5690  \\
	\bottomrule
\end{tabular}
\label{tab:cub}
\end{table}

\begin{table}[t]
\small
\setlength{\tabcolsep}{12pt}
\centering
\caption{The performance comparison of different loss functions on Dense-Haze \cite{Dense-Haze_2019} dataset with MSBDN \cite{dong2020multi} as baseline. We train the baseline from scratch and choose the best performance in the first 40K iterations.}
\begin{tabular}{lccc}
	\toprule
	\multirow{1}{*}{Label} 
	& LPIPS$\downarrow$ & PSNR$\uparrow$ & SSIM$\uparrow$  \\
	
	\midrule
        L1 Loss           & 0.4948 & 15.228 & 0.4974   \\
        Perceptual Loss   & 0.4921 & 15.609 & 0.5011   \\
        Adversarial Loss  & 0.4848 & 15.593 & 0.4981  \\
	\rowcolor{lightgray!40}DiffLoss     & 0.4731  & 15.682 & 0.5088  \\
	\bottomrule
\end{tabular}

\label{tab:loss}
\end{table}

\noindent 
{\bf Improvement on Image Classification.} 
We have demonstrated in \cref{fig:hTissue} that h-space possesses semantic attributes. Therefore, we utilize h-space as a loss to preserve the semantic information for low level tasks. We synthesized degraded images with low-light, haze, and rain using the origin CUB dataset. We train the model using common low-level datasets as mentioned before with and without DiffLoss, and use the trained model to restore the degraded CUB dataset. We choose to use Taylorformer, EfDeRain and IAT for different degradations. Finally, we employ a pre-trained VGG16\cite{simonyan2014very} and Resnet50\cite{he2016deep} networks to evaluate the accuracy of classification. From the \cref{tab:cub}, it is evident that with the assistance of our method, the model can preserve more semantic information after the restoration process. Please note that our focus is not on comparing with other methods, but rather on improving existing approaches.

\noindent 
{\bf Comparison with other loss functions.} 
Previous loss~\cite{deng2020hardgan,fu2021dw} that assist restoration networks have the following drawbacks:
(1) L1 or L2 loss works in pixel space, which may produce images deviated from natural distribution. (2) The VGG16 used in perceptual loss is pre-trained for high-level task, instead of low-level image restoration task. (3) Adversarial loss treats restoration network as generator and inserts an additional discriminator network and needs to train a discriminator for every restoration dataset, which is troublesome and time-consuming.
The wide distribution and mode convergence property enables diffusion model to be a powerful and general image prior, and fits for both general and specific low-level image restoration tasks. Besides, h-space in diffusion model is also found to reflect the semantics of images. As shown in \cref{fig:reshaze} and \cref{tab:loss}, our DiffLoss demonstrates excellent performance compared to other approaches. 

By pulling the intermediate sampling stages and h-space closer to that of clear images and leveraging the distribution sampling property of diffusion model, the restored results can be optimized to be more natural and recognition-aware, which is difficult to achieve for the previous methods.

\begin{figure}[t]
	\begin{center}
		\includegraphics[width=0.8\linewidth]{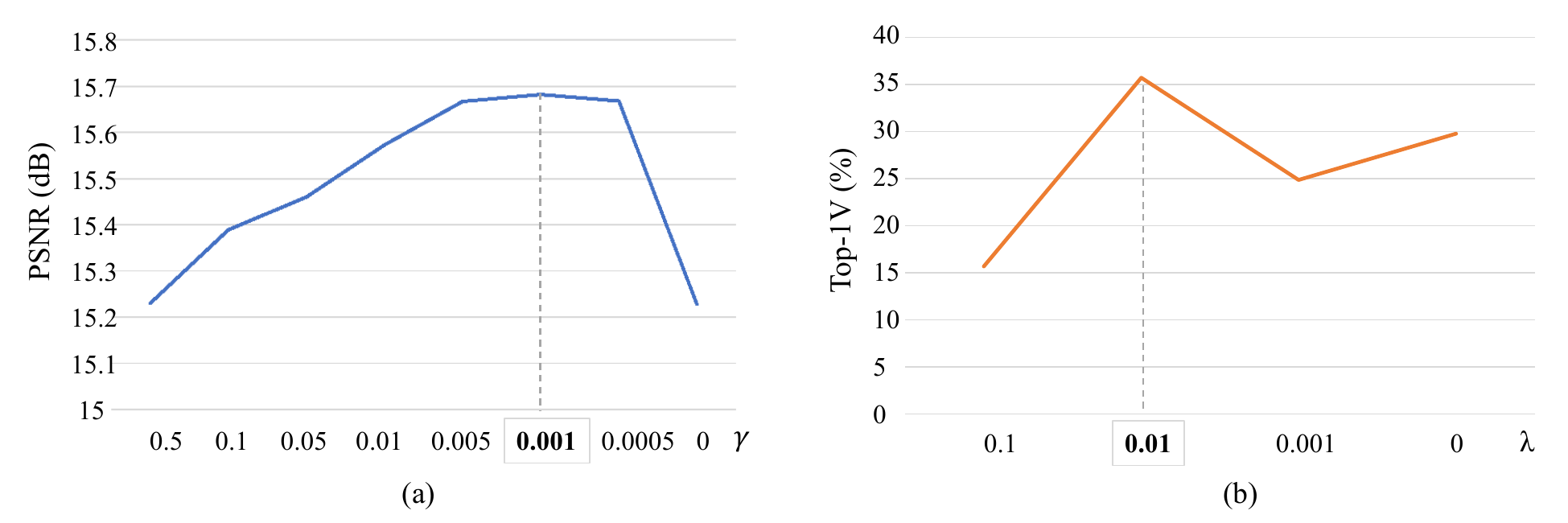}
	\end{center}
	\caption{Graph of PSNR and Top-1V(\%) with different weight of DiffLoss during training process on Dense-Haze \cite{Dense-Haze_2019} dataset with MSBDN \cite{dong2020multi} as baseline. We train the baseline from scratch and choose the best performance in the first 40K iterations.}
	\label{weight}
\end{figure}

\begin{table}[t]
\small
\setlength{\tabcolsep}{12pt}
\centering
\caption{The results of ablated models on Dense-Haze \cite{Dense-Haze_2019} dataset with GridNet~\cite{liu2019griddehazenet} as baseline.}
\begin{tabular}{lcccc}
	\toprule
	\multirow{1}{*}{Label} 
	& FID$\downarrow$  & PSNR$\uparrow$ & SSIM$\uparrow$  \\
	
	\midrule
        a  & 429.73  & 14.010 & 0.3681   \\
        b  & 343.31 & 14.824 & 0.4429   \\
	\rowcolor{lightgray!40} c (\textbf{ours})  & 293.01  & 14.907 & 0.4666  \\
	\bottomrule
\end{tabular}
\label{tab:ablation}
\end{table}
\subsection{Ablation study}
\label{sec:abl}
In this section, we perform several ablation studies to analyze the effectiveness of the proposed DiffLoss on Dense-Haze dataset. The studies include the following ablated models:
(a) With constraints on $\hat{x}_0$ only. 
(b) With constraints on $\hat{x}_{t-1}$ only. 
(c) Ours (final setting).
These models are trained using the same training setting as our method. The performance of these models is summarized in Table~\ref{tab:ablation}. Obviously, every design component in DiffLoss can elevate the performance.


 We also experiment with different loss value to get the optimal one. As shown in \cref{weight}, we experiment with different weight of DiffLoss and plot the PSNR-Loss weight curve. DiffLoss gets the optimal performance with loss weight $\gamma \in [0.0005, 0.005]$. In our method, we choose the loss weight of DiffLoss to be 0.001. This experiment is conducted with $\lambda=0$. For the optimal choice of $\lambda$, we find that when $\lambda=0.01$, it achieves the best performance. Note that in the experiments for selecting $\lambda$, we set $\gamma=0.001$.



\section{Conclusion}
   In this paper, we propose a new perspective from the correlation of degraded and natural image distribution that achieves effective image restoration. To achieve this, inspired by the powerful capability of the diffusion model for natural image sampling and generation, we embed the pre-trained diffusion model into restoration network as a auxiliary training mechanism to empower the learning capability and semantic attributes of neural networks for effective naturalness image restoration. 
   By equipping existing restoration networks with the DiffLoss in the training stage, we can substantially elevate their performance and yield more natural and semantic-aware restored images without involving additional computations in the inference stage.

\bibliographystyle{splncs04}
\bibliography{main}

\begin{thebibliography}{10}
\providecommand{\url}[1]{\texttt{#1}}
\providecommand{\urlprefix}{URL }
\providecommand{\doi}[1]{https://doi.org/#1}

\bibitem{Dense-Haze_2019}
Ancuti, C.O., Ancuti, C., Sbert, M., Timofte, R.: Dense haze: A benchmark for
  image dehazing with dense-haze and haze-free images. In: Proceedings of the
  IEEE International Conference on Image Processing. pp. 1014--1018 (2019)

\bibitem{ancuti2020nh}
Ancuti, C.O., Ancuti, C., Timofte, R.: {NH-HAZE}: An image dehazing benchmark
  with non-homogeneous hazy and haze-free images. In: Proceedings of the
  IEEE/CVF Conference on Computer Vision and Pattern Recognition Workshops. pp.
  444--445 (2020)

\bibitem{baranchuk2021label}
Baranchuk, D., Rubachev, I., Voynov, A., Khrulkov, V., Babenko, A.:
  Label-efficient semantic segmentation with diffusion models. arXiv preprint
  arXiv:2112.03126  (2021)

\bibitem{berman2016non}
Berman, D., Avidan, S., et~al.: Non-local image dehazing. In: Proceedings of
  the IEEE Conference on Computer Vision and Pattern Recognition. pp.
  1674--1682 (2016)

\bibitem{chakrabarti2016neural}
Chakrabarti, A.: A neural approach to blind motion deblurring. In: Computer
  Vision--ECCV 2016: 14th European Conference, Amsterdam, The Netherlands,
  October 11-14, 2016, Proceedings, Part III 14. pp. 221--235. Springer (2016)

\bibitem{chen2021hinet}
Chen, L., Lu, X., Zhang, J., Chu, X., Chen, C.: {HINet}: Half instance
  normalization network for image restoration. In: Proceedings of the IEEE/CVF
  Conference on Computer Vision and Pattern Recognition. pp. 182--192 (2021)

\bibitem{chen2019pms}
Chen, W.T., Ding, J.J., Kuo, S.Y.: Pms-net: Robust haze removal based on patch
  map for single images. In: Proceedings of the IEEE/CVF conference on computer
  vision and pattern recognition. pp. 11681--11689 (2019)

\bibitem{chen2018domain}
Chen, Y., Li, W., Sakaridis, C., Dai, D., Van~Gool, L.: Domain adaptive faster
  r-cnn for object detection in the wild. In: Proceedings of the IEEE
  Conference on Computer Vision and Pattern Recognition. pp. 3339--3348 (2018)

\bibitem{cho2021rethinking}
Cho, S.J., Ji, S.W., Hong, J.P., Jung, S.W., Ko, S.J.: Rethinking
  coarse-to-fine approach in single image deblurring. In: Proceedings of the
  IEEE/CVF international conference on computer vision. pp. 4641--4650 (2021)

\bibitem{choi2021ilvr}
Choi, J., Kim, S., Jeong, Y., Gwon, Y., Yoon, S.: Ilvr: Conditioning method for
  denoising diffusion probabilistic models. arXiv preprint arXiv:2108.02938
  (2021)

\bibitem{cui2022you}
Cui, Z., Li, K., Gu, L., Su, S., Gao, P., Jiang, Z., Qiao, Y., Harada, T.: You
  only need 90k parameters to adapt light: a light weight transformer for image
  enhancement and exposure correction. In: BMVC. p.~238 (2022)

\bibitem{deng2020hardgan}
Deng, Q., Huang, Z., Tsai, C.C., Lin, C.W.: {HardGAN}: A haze-aware
  representation distillation gan for single image dehazing. In: Proceedings of
  the European Conference on Computer Vision. pp. 722--738. Springer (2020)

\bibitem{dhariwal2021diffusion}
Dhariwal, P., Nichol, A.: Diffusion models beat gans on image synthesis.
  Advances in Neural Information Processing Systems  \textbf{34},  8780--8794
  (2021)

\bibitem{dong2020multi}
Dong, H., Pan, J., Xiang, L., Hu, Z., Zhang, X., Wang, F., Yang, M.H.:
  Multi-scale boosted dehazing network with dense feature fusion. In:
  Proceedings of the IEEE/CVF Conference on Computer Vision and Pattern
  Recognition. pp. 2157--2167 (2020)

\bibitem{fattal2008single}
Fattal, R.: Single image dehazing. ACM Transactions on Graphics (TOG)
  \textbf{27}(3), ~1--9 (2008)

\bibitem{fattal2014dehazing}
Fattal, R.: Dehazing using color-lines. ACM Transactions on Graphics (TOG)
  \textbf{34}(1),  1--14 (2014)

\bibitem{fu2021dw}
Fu, M., Liu, H., Yu, Y., Chen, J., Wang, K.: {DW-GAN}: A discrete wavelet
  transform gan for nonhomogeneous dehazing. In: Proceedings of the IEEE/CVF
  Conference on Computer Vision and Pattern Recognition. pp. 203--212 (2021)

\bibitem{fu2017clearing}
Fu, X., Huang, J., Ding, X., Liao, Y., Paisley, J.: Clearing the skies: A deep
  network architecture for single-image rain removal. IEEE Transactions on
  Image Processing  \textbf{26}(6),  2944--2956 (2017)

\bibitem{fu2017removing}
Fu, X., Huang, J., Zeng, D., Huang, Y., Ding, X., Paisley, J.: Removing rain
  from single images via a deep detail network. In: Proceedings of the IEEE
  conference on computer vision and pattern recognition. pp. 3855--3863 (2017)

\bibitem{gao2019dynamic}
Gao, H., Tao, X., Shen, X., Jia, J.: Dynamic scene deblurring with parameter
  selective sharing and nested skip connections. In: Proceedings of the
  IEEE/CVF conference on computer vision and pattern recognition. pp.
  3848--3856 (2019)

\bibitem{guo2022image}
Guo, C.L., Yan, Q., Anwar, S., Cong, R., Ren, W., Li, C.: Image dehazing
  transformer with transmission-aware 3d position embedding. In: Proceedings of
  the IEEE/CVF Conference on Computer Vision and Pattern Recognition. pp.
  5812--5820 (2022)

\bibitem{guo2021efficientderain}
Guo, Q., Sun, J., Juefei-Xu, F., Ma, L., Xie, X., Feng, W., Liu, Y., Zhao, J.:
  Efficientderain: Learning pixel-wise dilation filtering for high-efficiency
  single-image deraining. In: Proceedings of the AAAI Conference on Artificial
  Intelligence. vol.~35, pp. 1487--1495 (2021)

\bibitem{haas2023discovering}
Haas, R., Huberman-Spiegelglas, I., Mulayoff, R., Michaeli, T.: Discovering
  interpretable directions in the semantic latent space of diffusion models.
  arXiv preprint arXiv:2303.11073  \textbf{3}(6) (2023)

\bibitem{he2010single}
He, K., Sun, J., Tang, X.: Single image haze removal using dark channel prior.
  IEEE Transactions on Pattern Analysis and Machine Intelligence
  \textbf{33}(12),  2341--2353 (2010)

\bibitem{he2016deep}
He, K., Zhang, X., Ren, S., Sun, J.: Deep residual learning for image
  recognition. In: Proceedings of the IEEE Conference on Computer Vision and
  Pattern Recognition. pp. 770--778 (2016)

\bibitem{hendrycks2019benchmarking}
Hendrycks, D., Dietterich, T.: Benchmarking neural network robustness to common
  corruptions and perturbations. arXiv preprint arXiv:1903.12261  (2019)

\bibitem{jeong2024training}
Jeong, J., Kwon, M., Uh, Y.: Training-free content injection using h-space in
  diffusion models. In: Proceedings of the IEEE/CVF Winter Conference on
  Applications of Computer Vision. pp. 5151--5161 (2024)

\bibitem{jiang2021enlightengan}
Jiang, Y., Gong, X., Liu, D., Cheng, Y., Fang, C., Shen, X., Yang, J., Zhou,
  P., Wang, Z.: Enlightengan: Deep light enhancement without paired
  supervision. IEEE transactions on image processing  \textbf{30},  2340--2349
  (2021)

\bibitem{kawar2022denoising}
Kawar, B., Elad, M., Ermon, S., Song, J.: Denoising diffusion restoration
  models. arXiv preprint arXiv:2201.11793  (2022)

\bibitem{kim2021quality}
Kim, I., Han, S., Baek, J.w., Park, S.J., Han, J.J., Shin, J.: Quality-agnostic
  image recognition via invertible decoder. In: Proceedings of the IEEE/CVF
  Conference on Computer Vision and Pattern Recognition. pp. 12257--12266
  (2021)

\bibitem{kingma2013auto}
Kingma, D.P., Welling, M.: Auto-encoding variational bayes. arXiv preprint
  arXiv:1312.6114  (2013)

\bibitem{kupyn2018deblurgan}
Kupyn, O., Budzan, V., Mykhailych, M., Mishkin, D., Matas, J.: Deblurgan: Blind
  motion deblurring using conditional adversarial networks. In: Proceedings of
  the IEEE conference on computer vision and pattern recognition. pp.
  8183--8192 (2018)

\bibitem{ledig2017photo}
Ledig, C., Theis, L., Husz{\'a}r, F., Caballero, J., Cunningham, A., Acosta,
  A., Aitken, A., Tejani, A., Totz, J., Wang, Z., et~al.: Photo-realistic
  single image super-resolution using a generative adversarial network. In:
  Proceedings of the IEEE conference on computer vision and pattern
  recognition. pp. 4681--4690 (2017)

\bibitem{li2017aod}
Li, B., Peng, X., Wang, Z., Xu, J., Feng, D.: {AOD-Net}: All-in-one dehazing
  network. In: Proceedings of the IEEE International Conference on Computer
  Vision. pp. 4770--4778 (2017)

\bibitem{li2018end}
Li, B., Peng, X., Wang, Z., Xu, J., Feng, D.: End-to-end united video dehazing
  and detection. In: Proceedings of the AAAI Conference on Artificial
  Intelligence. vol.~32 (2018)

\bibitem{li2020zero}
Li, B., Gou, Y., Liu, J.Z., Zhu, H., Zhou, J.T., Peng, X.: Zero-shot image
  dehazing. IEEE Transactions on Image Processing  \textbf{29},  8457--8466
  (2020)

\bibitem{li2018recurrent}
Li, X., Wu, J., Lin, Z., Liu, H., Zha, H.: Recurrent squeeze-and-excitation
  context aggregation net for single image deraining. In: Proceedings of the
  European conference on computer vision (ECCV). pp. 254--269 (2018)

\bibitem{lim2017enhanced}
Lim, B., Son, S., Kim, H., Nah, S., Mu~Lee, K.: Enhanced deep residual networks
  for single image super-resolution. In: Proceedings of the IEEE conference on
  computer vision and pattern recognition workshops. pp. 136--144 (2017)

\bibitem{liu2019griddehazenet}
Liu, X., Ma, Y., Shi, Z., Chen, J.: {GridDehazeNet}: Attention-based
  multi-scale network for image dehazing. In: Proceedings of the IEEE/CVF
  International Conference on Computer Vision. pp. 7314--7323 (2019)

\bibitem{lugmayr2022repaint}
Lugmayr, A., Danelljan, M., Romero, A., Yu, F., Timofte, R., Van~Gool, L.:
  Repaint: Inpainting using denoising diffusion probabilistic models. In:
  Proceedings of the IEEE/CVF Conference on Computer Vision and Pattern
  Recognition. pp. 11461--11471 (2022)

\bibitem{luo2015removing}
Luo, Y., Xu, Y., Ji, H.: Removing rain from a single image via discriminative
  sparse coding. In: Proceedings of the IEEE international conference on
  computer vision. pp. 3397--3405 (2015)

\bibitem{mei2022bi}
Mei, K., Nair, N.G., Patel, V.M.: Bi-noising diffusion: Towards conditional
  diffusion models with generative restoration priors. arXiv preprint
  arXiv:2212.07352  (2022)

\bibitem{moran2020deeplpf}
Moran, S., Marza, P., McDonagh, S., Parisot, S., Slabaugh, G.: Deeplpf: Deep
  local parametric filters for image enhancement. In: Proceedings of the
  IEEE/CVF conference on computer vision and pattern recognition. pp.
  12826--12835 (2020)

\bibitem{ozdenizci2022restoring}
{\"O}zdenizci, O., Legenstein, R.: Restoring vision in adverse weather
  conditions with patch-based denoising diffusion models. arXiv preprint
  arXiv:2207.14626  (2022)

\bibitem{qian2018attentive}
Qian, R., Tan, R.T., Yang, W., Su, J., Liu, J.: Attentive generative
  adversarial network for raindrop removal from a single image. In: Proceedings
  of the IEEE conference on computer vision and pattern recognition. pp.
  2482--2491 (2018)

\bibitem{qin2020ffa}
Qin, X., Wang, Z., Bai, Y., Xie, X., Jia, H.: {FFA-Net}: Feature fusion
  attention network for single image dehazing. In: Proceedings of the AAAI
  Conference on Artificial Intelligence. vol.~34, pp. 11908--11915 (2020)

\bibitem{qiu2023mb}
Qiu, Y., Zhang, K., Wang, C., Luo, W., Li, H., Jin, Z.: Mb-taylorformer:
  Multi-branch efficient transformer expanded by taylor formula for image
  dehazing. In: Proceedings of the IEEE/CVF International Conference on
  Computer Vision. pp. 12802--12813 (2023)

\bibitem{ren2019progressive}
Ren, D., Zuo, W., Hu, Q., Zhu, P., Meng, D.: Progressive image deraining
  networks: A better and simpler baseline. In: Proceedings of the IEEE/CVF
  conference on computer vision and pattern recognition. pp. 3937--3946 (2019)

\bibitem{ren2021deblurring}
Ren, W., Zhang, J., Pan, J., Liu, S., Ren, J.S., Du, J., Cao, X., Yang, M.H.:
  Deblurring dynamic scenes via spatially varying recurrent neural networks.
  IEEE transactions on pattern analysis and machine intelligence
  \textbf{44}(8),  3974--3987 (2021)

\bibitem{rombach2022high}
Rombach, R., Blattmann, A., Lorenz, D., Esser, P., Ommer, B.: High-resolution
  image synthesis with latent diffusion models. In: Proceedings of the IEEE/CVF
  Conference on Computer Vision and Pattern Recognition. pp. 10684--10695
  (2022)

\bibitem{saharia2022image}
Saharia, C., Ho, J., Chan, W., Salimans, T., Fleet, D.J., Norouzi, M.: Image
  super-resolution via iterative refinement. IEEE Transactions on Pattern
  Analysis and Machine Intelligence  (2022)

\bibitem{sakaridis2018model}
Sakaridis, C., Dai, D., Hecker, S., Van~Gool, L.: Model adaptation with
  synthetic and real data for semantic dense foggy scene understanding. In:
  Proceedings of the European Conference on Computer Vision. pp. 687--704
  (2018)

\bibitem{sakaridis2018semantic}
Sakaridis, C., Dai, D., Van~Gool, L.: Semantic foggy scene understanding with
  synthetic data. International Journal of Computer Vision  \textbf{126}(9),
  973--992 (2018)

\bibitem{simonyan2014very}
Simonyan, K., Zisserman, A.: Very deep convolutional networks for large-scale
  image recognition. arXiv preprint arXiv:1409.1556  (2014)

\bibitem{songpseudoinverse}
Song, J., Vahdat, A., Mardani, M., Kautz, J.: Pseudoinverse-guided diffusion
  models for inverse problems. In: International Conference on Learning
  Representations

\bibitem{wah2011caltech}
Wah, C., Branson, S., Welinder, P., Perona, P., Belongie, S.: The caltech-ucsd
  birds-200-2011 dataset  (2011)

\bibitem{wang2020model}
Wang, H., Xie, Q., Zhao, Q., Meng, D.: A model-driven deep neural network for
  single image rain removal. In: Proceedings of the IEEE/CVF conference on
  computer vision and pattern recognition. pp. 3103--3112 (2020)

\bibitem{wang2022zero}
Wang, Y., Yu, J., Zhang, J.: Zero-shot image restoration using denoising
  diffusion null-space model. arXiv preprint arXiv:2212.00490  (2022)

\bibitem{wang2004image}
Wang, Z., Bovik, A.C., Sheikh, H.R., Simoncelli, E.P.: Image quality
  assessment: from error visibility to structural similarity. IEEE transactions
  on image processing  \textbf{13},  600--612 (2004)

\bibitem{wei1808deep}
Wei, C., Wang, W., Yang, W., Liu, J.: Deep retinex decomposition for low-light
  enhancement. arxiv 2018. arXiv preprint arXiv:1808.04560

\bibitem{wei2018deep}
Wei, C., Wang, W., Yang, W., Liu, J.: Deep retinex decomposition for low-light
  enhancement. arXiv preprint arXiv:1808.04560  (2018)

\bibitem{wei2023raindiffusion}
Wei, M., Shen, Y., Wang, Y., Xie, H., Wang, F.L.: Raindiffusion: When
  unsupervised learning meets diffusion models for real-world image deraining.
  arXiv preprint arXiv:2301.09430  (2023)

\bibitem{whang2022deblurring}
Whang, J., Delbracio, M., Talebi, H., Saharia, C., Dimakis, A.G., Milanfar, P.:
  Deblurring via stochastic refinement. In: Proceedings of the IEEE/CVF
  Conference on Computer Vision and Pattern Recognition. pp. 16293--16303
  (2022)

\bibitem{wu2021contrastive}
Wu, H., Qu, Y., Lin, S., Zhou, J., Qiao, R., Zhang, Z., Xie, Y., Ma, L.:
  Contrastive learning for compact single image dehazing. In: Proceedings of
  the IEEE/CVF Conference on Computer Vision and Pattern Recognition. pp.
  10551--10560 (2021)

\bibitem{xie2023semi}
Xie, S., Ma, Y., Xu, W., Qiu, S., Sun, Y.: Semi-supervised learning for
  low-light image enhancement by pseudo low-light image. In: 2023 16th
  International Congress on Image and Signal Processing, BioMedical Engineering
  and Informatics (CISP-BMEI). pp.~1--6. IEEE (2023)

\bibitem{yang2017deep}
Yang, W., Tan, R.T., Feng, J., Liu, J., Guo, Z., Yan, S.: Deep joint rain
  detection and removal from a single image. In: Proceedings of the IEEE
  conference on computer vision and pattern recognition. pp. 1357--1366 (2017)

\bibitem{yang2022self}
Yang, Y., Wang, C., Liu, R., Zhang, L., Guo, X., Tao, D.: Self-augmented
  unpaired image dehazing via density and depth decomposition. In: Proceedings
  of the IEEE/CVF Conference on Computer Vision and Pattern Recognition. pp.
  2037--2046 (2022)

\bibitem{yang2023visual}
Yang, Z., Huang, J., Chang, J., Zhou, M., Yu, H., Zhang, J., Zhao, F.: Visual
  recognition-driven image restoration for multiple degradation with intrinsic
  semantics recovery. In: Proceedings of the IEEE/CVF Conference on Computer
  Vision and Pattern Recognition. pp. 14059--14070 (2023)

\bibitem{yufrequency}
Yu, H., Zheng, N., Zhou, M., Huang, J., Xiao, Z., Zhao, F.: Frequency and
  spatial dual guidance for image dehazing

\bibitem{yu2021frechet}
Yu, Y., Zhang, W., Deng, Y.: Frechet inception distance (fid) for evaluating
  gans. China University of Mining Technology Beijing Graduate School
  \textbf{3} (2021)

\bibitem{zhang2018densely}
Zhang, H., Patel, V.M.: Densely connected pyramid dehazing network. In:
  Proceedings of the IEEE Conference on Computer Vision and Pattern
  Recognition. pp. 3194--3203 (2018)

\bibitem{zhang2018unreasonable}
Zhang, R., Isola, P., Efros, A.A., Shechtman, E., Wang, O.: The unreasonable
  effectiveness of deep features as a perceptual metric. In: Proceedings of the
  IEEE conference on computer vision and pattern recognition. pp. 586--595
  (2018)

\bibitem{zhang2021beyond}
Zhang, Y., Guo, X., Ma, J., Liu, W., Zhang, J.: Beyond brightening low-light
  images. International Journal of Computer Vision  \textbf{129},  1013--1037
  (2021)

\bibitem{zhang2019kindling}
Zhang, Y., Zhang, J., Guo, X.: Kindling the darkness: A practical low-light
  image enhancer. In: Proceedings of the 27th ACM international conference on
  multimedia. pp. 1632--1640 (2019)

\bibitem{zhang2018image}
Zhang, Y., Li, K., Li, K., Wang, L., Zhong, B., Fu, Y.: Image super-resolution
  using very deep residual channel attention networks. In: Proceedings of the
  European conference on computer vision (ECCV). pp. 286--301 (2018)

\bibitem{zhang2018residual}
Zhang, Y., Tian, Y., Kong, Y., Zhong, B., Fu, Y.: Residual dense network for
  image super-resolution. In: Proceedings of the IEEE conference on computer
  vision and pattern recognition. pp. 2472--2481 (2018)

\bibitem{zhang2023unsupervised}
Zhang, Z., Lin, L.L., Zhu, Y., Zhao, Z., et~al.: Unsupervised discovery of
  interpretable directions in h-space of pre-trained diffusion models. arXiv
  preprint arXiv:2310.09912  (2023)

\bibitem{zhu2020zero}
Zhu, A., Zhang, L., Shen, Y., Ma, Y., Zhao, S., Zhou, Y.: Zero-shot restoration
  of underexposed images via robust retinex decomposition. In: 2020 IEEE
  International Conference on Multimedia and Expo (ICME). pp.~1--6. IEEE (2020)

\end{thebibliography}
\end{document}